\documentclass{article}

\usepackage[final]{nips_2017}

\usepackage{times}
\usepackage{latexsym}
\usepackage[utf8]{inputenc} 
\usepackage[T1]{fontenc}    
\usepackage{hyperref}       
\usepackage{url}            
\usepackage{booktabs}       
\usepackage{amsfonts}       
\usepackage{nicefrac}       
\usepackage{microtype}      
\usepackage{times}
\usepackage{amsmath}
\usepackage{amssymb}
\usepackage{graphicx}
\usepackage{adjustbox}
\usepackage{wrapfig}
\usepackage{array}
\usepackage{color,soul}
\usepackage{multicol}
\usepackage{multirow}
\usepackage{enumitem}
\usepackage{subcaption}
\usepackage{natbib}
\usepackage{algorithm}
\usepackage{algorithmic}
\usepackage{footmisc}
\usepackage{hyperref}
\usepackage{xspace}

\usepackage{tikz}
\usepackage{pgfplots}
\usepgfplotslibrary{fillbetween}
\definecolor{red1}{RGB}{157, 66, 83}
\definecolor{green1}{RGB}{63, 127, 67}
\definecolor{lightgreen}{RGB}{179,225,172}
\definecolor{lightred}{RGB}{228,160,136}
\definecolor{lightblue}{RGB}{176,226,255}
\definecolor{green2}{RGB}{181,213,167}
\definecolor{green3}{RGB}{244,177,131}
\definecolor{yellow2}{RGB}{255,255,153}

\pgfplotsset{compat=1.6}
\usepackage{subcaption} 

\makeatletter
\usepackage{tikz}
\usetikzlibrary{arrows,chains,matrix,positioning,scopes}
\makeatletter
\tikzset{join/.code=\tikzset{after node path={%
\ifx\tikzchainprevious\pgfutil@empty\else(\tikzchainprevious)%
edge[every join]#1(\tikzchaincurrent)\fi}}}
\makeatother
\tikzset{>=stealth',every on chain/.append style={join},
         every join/.style={->}}
\tikzstyle{labeled}=[execute at begin node=$\scriptstyle,
   execute at end node=$]

\usepackage{hyperref}

\title{Scaffolding Networks: Incremental Learning and Teaching Through Questioning}
\author{Asli Celikyilmaz, Li Deng, Lihong Li, Chong Wang\\ Microsoft Research
}

\begin{document}

\maketitle
\begin{abstract}
We introduce a new paradigm of learning for reasoning, understanding, and prediction, as well as the scaffolding network to implement this paradigm. The scaffolding network embodies an incremental learning approach that is formulated as a teacher-student network architecture to teach machines how to understand text and do reasoning. The key to our computational scaffolding approach is the interactions between the teacher and the student through sequential questioning. The student observes each sentence in the text incrementally, and it uses an attention-based neural net to discover and register the key information in relation to its current memory. Meanwhile, the teacher asks questions about the observed text, and the student network gets rewarded by correctly answering these questions. 
The entire network is updated continually using reinforcement learning. 
Our experimental results on synthetic and real datasets show that the scaffolding network not only outperforms state-of-the-art methods but also learns to do reasoning in a scalable way even with little human generated input.
\end{abstract} 
\section{Introduction}
\label{intro}
Today's machines are mostly designed to understand human language from written or spoken text. 
Most related work has used fixed training data of labeled text focusing on learning
lower-level semantics in sentences (such as semantic tagging or parsing), later to be expanded to higher-level semantics in documents or dialogs (such as sentiment detection, question-answering, machine translation or task completion dialogs). 
Nevertheless, real human learning is incremental, repetitive, compositional, and is enhanced through interaction with a teacher or other humans. New and exciting work has recently started to emerge that use human-like learning, e.g., dialog-based language understanding~\cite{WestonDialog}, interactive language learning~\cite{Wang16}, compositional learning~\cite{Andreas,Baidu}, to name a few. 

In this work, we explore teaching machines to learn concepts as conveyed in text with a new architecture that uses an incremental learning framework.
We draw inspiration from real-life instructional scaffolding teaching \cite{Youb,Hogan}, a process through which a teacher adds support for students to enhance learning. 
A common method in scaffolding teaching is to repeatedly question students to check for understanding, which 
systematically builds on students’ experiences and knowledge.  
We present a neural network architecture that imitates scaffolding teaching for understanding text and do reasoning. 

The Scaffolding Network (SN) 
is comprised of a teacher and a student sharing the same environment, and combines several features into a single, effective network. The student observes text one sentence at a time, which contains limited or partial information about the state of the world; no prior information of entities, slots or relations between them is provided.
Its goal is to perform sentence-level interpretation, using a recurrent memory structure and attention mechanism to track the changes of facts over the course of story or dialog, and identify new ones that has not been observed before.
Meanwhile, the teacher, being a companion, generates question-answer pairs related to the text observed so far and poses these questions to the student.
Observing a sentence and a question, the student consequently learns to encode and answer the question based on the reward signal it receives from the environment. 
We adopt a reinforcement learning (RL) framework which provides an ideal setting for step by step teaching of semantic understanding. The student aims to maximize its rewards that it obtains upon answering generated questions. 
Our contributions are:
\vspace{-0.06in}
\begin{itemize}[leftmargin=0.2in]
\item We present a neural network model, which introduces a teacher-student framework to \textit{incrementally learn} and \textit{reason} about the state of the world by extracting and registering the facts that are partially observable. Our models are scalable due to incremental encoding of each sentence.
\vspace{-0.05in}
\item We introduce a teacher that can generate scaffolding question-answer pairs and guides the student to accurately estimate the relations between facts and their changing states. 
%
%
\vspace{-0.05in}
\item We evaluate the usefulness of our approach against baseline supervised learning models on story-based reasoning and dialog tasks. We show promising improvements on reasoning about objects and their relations, especially when we reduce the human generated input.
\end{itemize}

\section{Related Work}
We draw particular inspiration for scaffolding networks from Weston's dialog-based language learning approach, which gets feedback from a teacher and various other signals while the student imitates such feedback in order to learn to answer questions~\cite{WestonDialog}. 
Li et al. presents a conversational agent that gets feedback from a human teacher based on its response to improve question answering skills~\cite{Jiwei}.
Another recent work~\cite{GuoICLR17} uses dialog-based interactive learning in which an agent iteratively asks questions about relevant missing facts, so it learns to correctly answer a target question. 
Several memory networks expansions have also shown promising improvements on dialog datasets \cite{Bordes2016,Seo,GE2EMN}.
Unlike these work, which mainly search for clues about a particular question, our goal is to learn to encode the entire text incrementally to do reasoning and learn to answer as many questions as possible. Also, the key to our approach is that the teacher helps student focus on the state changes of entities and register into the episode memory, which was not fully explored in these work.

Another inspiration of this work is from the incremental learning framework in which the input text is received in sequences of sentences and encoded separately.  
While \cite{Yelong} uses an Markov Decision Process (MDP) framework to learn when to stop reading to answer a focused question, \cite{Graber} builds an incremental classifier for a trivia game to decide whether additional features are needed. The closest to our work is the Recurrent Entity Networks \cite{WestonICLR17}, in which the authors try to learn an internal state representation of each sentence in a sequential order and store in memory by parallel recurrent units with tied weights using gating functions. Although our network's structure has similarities, they use supervised objective while we view the task as an MDP~\cite{Zubek}, in which our agent choses its actions to maximize its rewards towards learning to encode text by answering teacher's questions.  
\section{The Scaffolding Networks}
\label{SN}
\begin{wrapfigure}{r}{0.5\textwidth}
\vskip -0.3in
\begin{center} 
\adjustbox{trim={0.02\width} {0.18\height} {0.03\width} {0.02\height},clip}%
{
\includegraphics[width=0.5\textwidth]{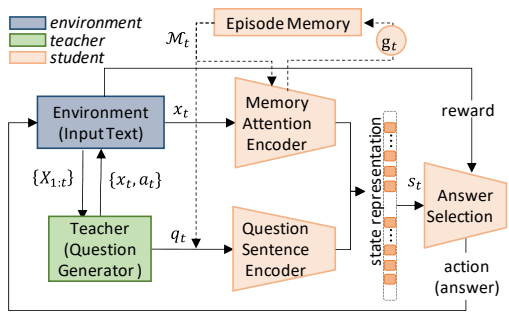}
}
\end{center} 
\vskip -0.15in
\caption{\small Illustration of scaffolding networks.}
\label{fpic}
\vskip -0.15in
\end{wrapfigure}
The scaffolding networks (Figure 1) comprise of a \textit{teacher} and a \textit{student} sharing an \textit{environment}. 
The environment provides the student and the teacher the sentences from input text one at a time. 
The teacher component generates scaffolding question-answer pairs about the input text observed so far.  
Contrary to the teacher-student framework literature \cite{Suay,Zimmer}, our teacher is not a RL agent that directly influences student in taking actions.
Our teacher guides the student to \textit{focus} on parts of the text which might be harder for the student to extract information. 

The student observes a sentence from the environment and a question from the teacher at time $t$. 
It starts in an initial state (no memory), encodes the sentence together with its current memory using attention network, thereby enabling reasoning over entailments of pairs of memory cells. 
This way the student identifies new information to register into memory.
Meanwhile, it processes the sentence together with the posed question to capture similarities.
It aims to maximize the sum of all rewards until the end of the document is reached by correctly answering as many questions as possible in an episode through trial and error via RL algorithm. 
Depending on student's answer (the action), the environment moves on to provide the next sentence (if the answer is correct), or stays on providing the same sentence until the student learns to answer one question correctly (or until maximum trials are exhausted), while the teacher generates a new question each time. 
We denote the \textit{episode} as a sequence of time steps from an initial
state (the first sentence 
in text) to a terminal state (the end of document is reached or as set by the environment). 
The input to the scaffolding network is an input text, $\mathcal{X}$, of a sequence of sentences: $\mathcal{X}=(x_1,\ldots,x_T)$ denoting the environment from which the student receives sentences incrementally during the course of the episode.
Each sentence at time $t$ is a sequence of $n$ words, $x_t=(w_{t,1},\ldots,w_{t,n})$. 

\subsection{The Student}
The \textit{memory attention encoder} of the student in Figure~\ref{fpic} serves to   
(\textit{i}) identify whether there is new information in the current sentence in relation to its episode memory using \textit{soft attention}, and 
(\textit{ii}) control how much of its episode memory should be updated with this new information. The \textit{question-sentence encoder} serves to search for clues in the current sentence and the episode memory about the facts in the question, and the \textit{answer selection} module
learns a policy to answer questions posed by the teacher. 
The student observes a sentence $x_t$ at each time step of the episode.
Each token $w_{t,i}$ in sentence $x_t$ is represented with embedding vector, which are then processed through a long short-term memory (LSTM)~\cite{Hochreiter} model layer to obtain
hidden states
of each token $\mathbf{H}_t=(h_{t,1},\ldots,h_{t,n})$. 
\subsubsection{Memory Attention Encoder}
\label{AN}
\begin{figure}
\begin{minipage}[t]{\dimexpr.5\textwidth-1em}
\begin{center} 
\adjustbox{trim={0.05\width} {0.1\height} {0.03\width} {0.05\height},clip}%
{
\includegraphics[width=0.9\textwidth]{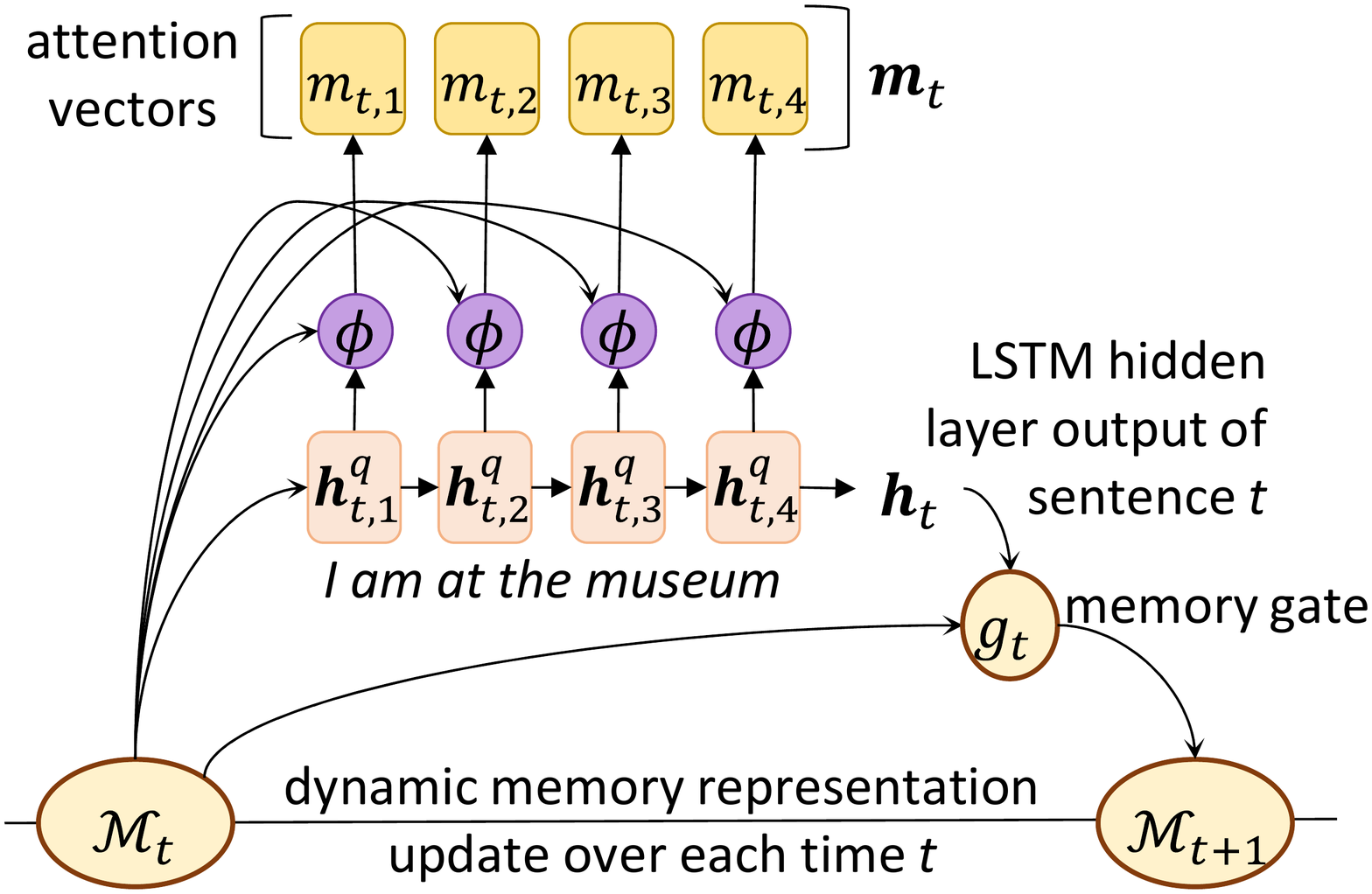}
}
\end{center} 
\vskip -0.15in
\caption{\small Memory Attention Encoder (from Figure 1) illustrating calculation of the soft attention vector and the gated episode memory update through time.}
\label{attention}
\vskip -0.15in
\end{minipage}\hfill
\begin{minipage}[t]{\dimexpr.5\textwidth-1em}
\begin{center} 
\adjustbox{trim={0.06\width} {0.2\height} {0.05\width} {0.02\height},clip}%
{
\includegraphics[width=0.9\linewidth]{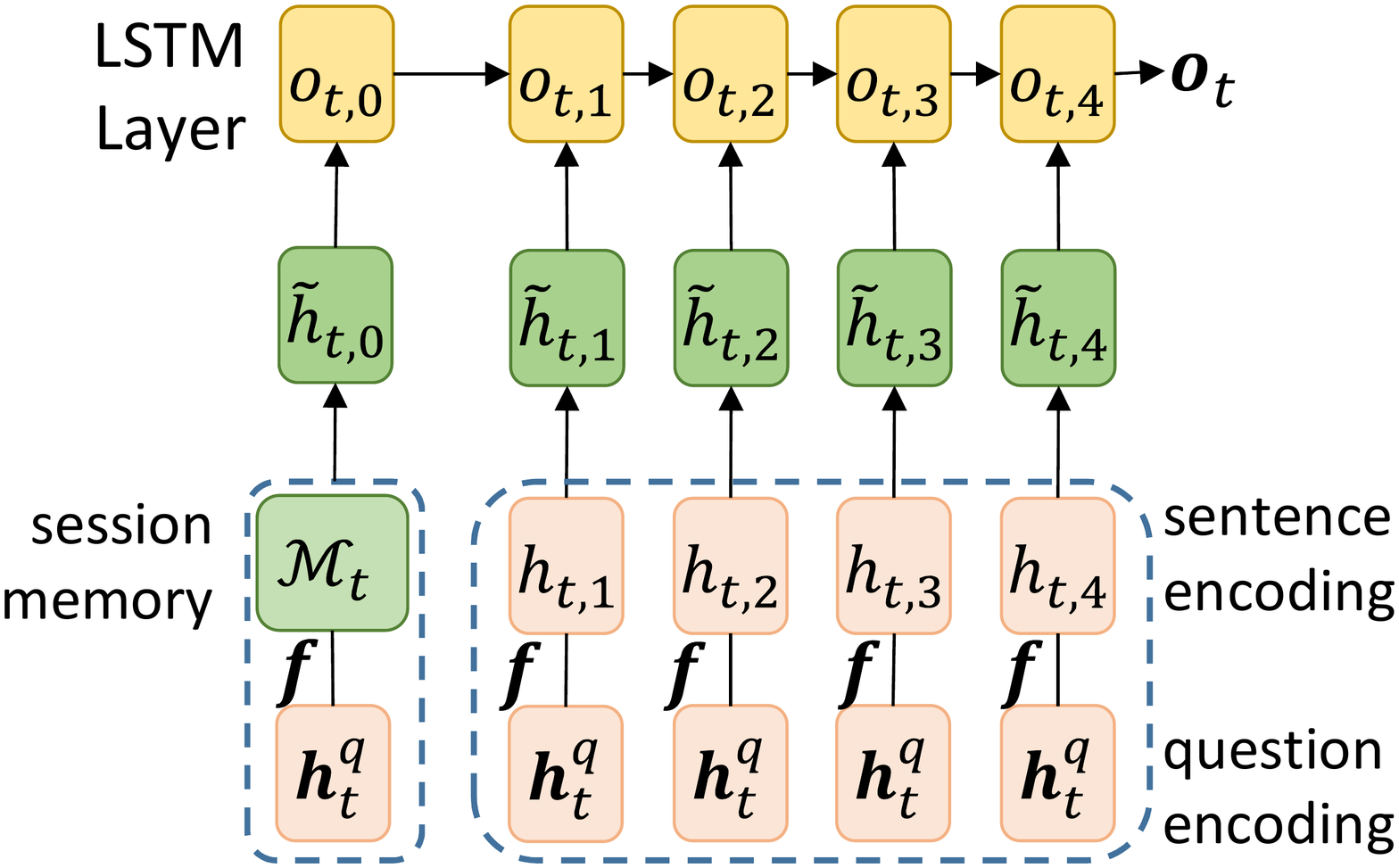}
}
\end{center} 
\vskip -0.15in
\caption{\small Question-Sentence Encoder (from Figure 1). $\mathcal{M}_t$ is the current episode memory, and $\mathbf{h}_t^q$ is the encoded question.}
\label{qs}
\vskip -0.15in
\end{minipage}
\end{figure}

\noindent\textbf{Soft Attention.} 
Rocktaschel et al. learns an entailment relation between two sentences (a premise and a hypothesis text) by word-by-word soft attention to encourage reasoning over entailment of pairs of words~\cite{Tim}. 
We learn a similar entailment between the observed sentence and the current memory at $t$, to determine if there is a new information in the observed sentence.
Weaker entailment (attention) indicates that the information in the current sentence compared to the memory is different (e.g., a new entity present in the current sentence but not in the memory). 
With the soft attention in Figure~\ref{attention}, we produce intermediate attention representations $\mathbf{m}_t$ as a non-linear combination of the episode memory $\mathcal{M}_{t}$ and the 
unrolled hidden states of the encoded sentence $h_{t,i}$ $\in$ $\mathbf{H}_t$:
\begin{equation}
\mathbf{m}_t \leftarrow \phi(W^{x}\mathbf{H}_t+W^{h}\mathcal{M}_{t}\otimes \mathbf{I}_n)\,,
\end{equation}
$W^{x}$ and $W^{h}$ are weight matrices to be learned, and the output product ($W^{h}\mathcal{M}_{t}\otimes \mathbf{I}_n$) repeats the memory representation $\mathcal{M}_{t}$ $n$ times.
$\phi$ is the activation function; in our experiments, we used the sigmoid activation function.
Each column $m_{t,i}$ of the output attention matrix $\mathbf{m}_t$ is the un-normalized attention of the $i$th word of sentence $x_t$. We concatenate each column to generate one soft attention vector $\mathbf{m}_t$ (overusing the notation), which partially forms the current state representation. 

\noindent\textbf{Episode Memory Update.} 
\label{dm}
The student determines how much of the episode memory should be updated with the information provided in the new sentence using a gating function (see Figure~\ref{attention}):
\begin{eqnarray}
g_t &\leftarrow& \phi(W^C\mathbf{h}_{t}+W^P\mathcal{M}_t)\\
\mathcal{M}_{t+1} &\leftarrow& \mathbf{h}_{t}+g_t \odot \mathcal{M}_{t}\,.
\label{gatingfunction2}
\end{eqnarray}
$W^C$ and $W^P$ are weight matrices to be learned.  They represent the weights associated with the current sentence and the episode memory at time $t$.
$\mathbf{h}_{t}$ is the last hidden state of the current sentence $x_t$ LSTM.
We choose hyperbolic tangent for the activation $\phi$. The way we look at the gating function $g_t$ is that, it allows some cells in the memory $\mathcal{M}_t$ to be forgotten and be replaced with the new ones $h_t$, as the information flows through time - as the dialog progresses or stories evolve.   
\subsubsection{Question-Sentence Encoder}
\label{QE}
At each time $t$, the teacher generates one or more questions which are concatenated into a single question $q_t$. 
Student maps the question $q_t$ into a vector form to generate its fixed length semantic representation.
Each word $a_{i}$ in the question $q_t=(a_1\cdots a_n)$
is encoded by mapping onto embedding layer and then processed through another LSTM layer to obtain the last hidden state vector $\mathbf{h}^q_t$. 
Given the fixed length representation of the sentence at time $t$ in semantic space $\mathbf{h_t}$, and the $\mathbf{h_t^q}$, the student applies a linear combination followed by another LSTM layer (Figure~\ref{qs}) as follows: 
\begin{eqnarray}
\mathbf{\tilde{h}}_{t,0} \leftarrow f(\mathcal{M}_{t},\mathbf{h}_t^q) \ \ \ \ \ \ ; \ \ \ \ \ \
\mathbf{\tilde{h}}_{t,i} \leftarrow f(\mathbf{h}_{t,i}+\mathbf{h}_t^q)  \ \ \ \ \ \ ; \ \ \ \ \ \
\mathbf{o}_t \leftarrow \mathrm{LSTM}(\mathbf{\tilde{h}}_{t,0},\mathbf{\tilde{h}}_t)
\end{eqnarray}
An interaction function $f(\cdot)$ is used to approximate the similarity between the question $\mathbf{h_t^q}$ and the memory $\mathcal{M}_t$ that gives the initial state $\mathbf{\tilde{h}}_{t,0}$ for the question-sentence LSTM. 
We merge the question $\mathbf{h_t^q}$ with the encoded sentence word vectors $h_{t,i}$$\in$$\mathbf{h}_t$ using the same interaction function to obtain $\tilde{h}_{t,i}$$\in$$\mathbf{\tilde{h}}_t$ approximating similarity between the current sentence and the question. 
The interaction function could be an inner product, or a nonlinear function such as a
deep neural network. In our experiments, both gave similar results.
For simplicity, in our experiments we mostly
use the inner product interaction function.
The last hidden state, $\mathbf{o}_t$, is used as part of the state representation.
\subsubsection{Answer Selection}
\label{scaffold}
\noindent\textbf{State.} 
At each time step $t$, the student agent concatenates the soft attention vector ($\mathbf{m}_t$), the output of the question-sentence encoder ($\mathbf{o}_t$) as well as the question representation ($\mathbf{h}_t^q$) to form the current state representation: $s_t=\left[\mathbf{o}_t;\mathbf{m}_t;\mathbf{h}_t^q\right]$. 

\noindent\textbf{Action.} 
An action corresponds to an answer, which is an entity from a task-specific entity set.
The agent selects an action $a_t$ at time $t$ from a finite set of actions $\mathcal{A}$=$\{1,\ldots,\mathcal{K}\}$, 
then collects a reward from the environment.  
The agent's action affects the next operations of the whole network, including whether the next sentence from input text is to be read, and how the teacher samples question-answer pairs (see $\S$~\ref{QN}); this affects the next state, $s_{t+1}$, observed by the student.
The teacher's strategy helps student explore different question-sentence pairs, yielding exploration of "novel" states.

\noindent\textbf{Reward.} 
We consider the case when the return is partially received at the intermediate steps in the form of rewards $r_t\in$\{+1,-1\} depending on answering questions correct or wrong.
If the answer is wrong, the student does not observe a new sentence, but rather the teacher generates a new question, and the student encodes the same sentence with the new question to form a new state. 
The student runs its policy and produces an answer and collects a new reward. This process continues until the student gets a correct answer or a maximum of trials are exhausted. 
The reward at terminal is a larger positive number which is proportional to the number of correctly answered questions given the episode length. The environment decides to terminate the episode if the student fails to provide correct answers to more than an empirically defined threshold. 

\noindent\textbf{Policy.}  
We learn the optimal policy $\pi^*$ through a Deep Q-Network (DQN) \cite{Mnih2015}.
The Q-function for the policy $\pi$ measures the expected cumulative discounted reward for each state-action pair $(s,a)$ if the agent chooses $a$ in state $s$ and then follows $\pi$ thereafter: $Q^{\pi}(s,a)=\mathbb{E}_{\pi} [ \sum_{i=0}^T\gamma^ir_{t+i}|s_t=s, a_t=a ]$, where $\gamma$ is the discount factor. 
During learning the optimal policy, $\epsilon$-greedy exploration is used with $\epsilon=0.1$ which decays over time.
We approximate the Q-function by a 2-layer neural network $Q(s,a;\theta)$, where $\theta$ are parameters to be learned.
As commonly done, we represent \textit{experiences} as tuples of transitions, ($(s_t,a_t,r_t,s_{t+1})\in\mathcal{D}_e$). 
We use \textit{experience reply} \cite{Mnih2015} by maintaining a buffer of experiences and training on randomly selected mini-batches. 
\subsection{The Teacher}
\label{QN}
The teacher, early in the training, serves to generate rather simple questions by sampling a sentence from so far observed sentences $\mathcal{X}_{1:t}$. 
When the student performance on a held-out set reaches a plateau, the teacher starts sampling multiple sentences from $\mathcal{X}_{1:t}$ to generate multiple questions having the same answer. 
The motivation is to ask questions that explicitly conveys spatially distant facts related to the answer and remind the student about such long distance dependencies (see Table~\ref{ex-app-log}). 

Every-time the student observes a sentence, it should not only track new facts, but also possible state changes about the facts.
For example, if the student observes a new sentence \textit{'the park is on my east'}, it should easily track that the '\textit{park}' is a new entity if it has not been observed before. 
However, if student have already observed the sentence \textit{'the museum is on the east of the park'}, then what it tracks should be the state change rather than the new fact because the new sentence indicates a location change. 
This may not be equally easy for the student to track with the entailment based memory attention encoder. 
We design the teacher to help student focus on such cases by
introducing a strategy called \textit{Importance Measure} $I(x_t)$ extending \cite{Zimmer}. 
The teacher first evaluates whether there is a state change in a given sentence as follows: 
At each time $t$, the student does forward pass and calculates the current sentence attention vector $\mathbf{\bar{m}}_t$ by taking the average of each of its word attention vectors $m_{t,i}$ : $\mathbf{\bar{m}}_t$=$\sum_i^n m_{t,i}$ from $i$=$1\cdots n$.
It also tracks the episode attention by measuring a moving average of average sentence attentions $\mathbf{\bar{m}}_{1:t}$=$ (\mathbf{\bar{m}}_{1:t-1}$+$\mathbf{\bar{m}}_t)$/2 over the episode. 
The teacher then measures the importance based on the similarity between the sentence $\mathbf{\bar{m}}_t$ and episode $\mathbf{\bar{m}}_{1:t}$ attentions:
\begin{eqnarray}
I(x_t) &\leftarrow& \textstyle \frac{\mathbf{\bar{m}}_{1:t}\cdot \mathbf{\bar{m}_t}}{ ||\mathbf{\bar{m}}_{1:t}||||\mathbf{\bar{m}_t}||}
\end{eqnarray}
We use linear kernel (cosine similarity) although any non-linear kernel can be used. When the average attention of the current sentence is similar to the episode's attention, it indicates that the current sentence may contain facts similar to so far observed sentences in the episode, hence a possible state change. The student should be questioned to verify that it captures this information correctly.

To generate question-answer pairs, our teacher uses the Question Generator (QG) in \cite{Heilman}\footnote{Code available in 
\url{http://www.cs.cmu.edu/~ark/mheilman/questions/}}, a rule-base system. 
Identifying various constructions such as conjunctions, subordinate clauses, and appositives, it generates multiple questions from a single sentence that conveys multiple pieces of information. 
QG uses Stanford Parser\cite{StanfordParser} to find sentence boundaries, tokenize, and parse the sentence and Tregex\cite{Tregex}, tree searching language, 
to identify various constructions (e.g., non-restrictive appositives). 
After identifying the key nodes with Tregex, QG manipulates
trees using the Stanford Parser, which allows for inserting and
deleting children, changing tree node labels, etc. to generate questions (see Table~\ref{ex-app-log}).

The teacher generates questions at each time $t$ using QG as follows: If the importance $I(x_t)$ of current sentence is greater than a given (empirically optimized) threshold, the teacher decides that the state of an entity might have changed.  
Using QG, it generates a set of question-answer pairs from current $x_t$. Additionally, it may also choose to select zero or more sentences from $\mathcal{X}_{1:t}$ to generate sets of question-answer pairs. Then it chooses one question from each set that has the same answer and forms question $q_t$ by concatenating each question.
If the importance is low, then it randomly chooses one or more sentences from $\mathcal{X}_{1:t}$ and uses QG to generate a question as explained above. 
\section{Experimental Setup}
\label{experiments}
\subsection{Training the Scaffolding Network}
If no validation data is provided, we withhold 10\% of the data for hyper-parameter tuning.
We use hidden size of 128 for all experiments. The weights in the input and output units were initialized with zero mean and standard deviation of 1/$\sqrt{d}$. 
L2 weight decay of 0.001 is used for all weights. 
All models are trained with ADAM~\cite{Kigma} with initial learning rates set by a grid search over $\{$0.1, 0.01$\}$.
We assign a reward of $\mp$1 depending on the answer being wrong or right when the question is posed by the teacher; or $\mp$(\textit{k}/T)*10 when the questions are corpus provided (at the end of the text) where \textit{k} is the correct answer count and T is the episode length. 
Following \cite{WestonNIPS2015}, we repeat each training 10 times with different random initializations and choose the best model based on validation performance and report test results of the best model. 

Our benchmark models use multi-hop/multi-layer memory networks and show promising improvements compared to single hop/layer counterparts; thus, we use 3-layer LSTMs for the question-sentence encoder for all experiments. For all scaffolding models, an episode is terminated and deemed unsuccessful when student fails to answer number of questions exceeding more than 50\% of sentences in the input text. 
\subsection{Datasets}
\label{data}
\noindent\textbf{Travel Log datasets.} This dataset is a collection of documents about travel narratives. Our goal in creating these datasets is to measure the ability of our scaffolding network in reasoning tasks which require learning relations between multiple facts and tracking their changing states.  
Each text is constructed by placing a traveler agent in a town defined by a 9$\times$9 grid, which is full of attractions. 
The agent starts randomly in one location and wanders around the town by randomly choosing a direction to the next move (i.e., \textit{north}, \textit{west}, \textit{south}, \textit{east}) and logs each move. As agent moves to a new direction, 
logs any attraction nearby with respect to the current location (e.g. \textit{there is a museum on my left.}) 
End of each log, a question is asked about the location of a randomly chosen attraction, (e.g.,\textit{what is north of the museum?}).
The task is to find the correct answer, which can only be inferred from the text by tracing the traveler's steps.
The complexity of the task increases as the number of attractions is increased.
Five different datasets are generated by varying the number of attractions, i.e., complexity of the reasoning task. 
Each dataset contains 1000 training and test logs.
(see Appendix-A) 

\noindent\textbf{bAbI Dialog datasets.} Our goal in using the dialog data is to test the capacity of the scaffolding networks to provide the correct response while tracking changing user goals. 
The dialog
dataset \cite{Bordes2016} is organized as set of 5 tasks (T1,..,T5) each focusing on a specific objective: issuing API
calls, updating API calls, displaying options, providing extra-information, conducting full
dialogs (the aggregation of the first 4 tasks). 
These data are synthetically generated based
on a knowledge base consisting of facts which define all the restaurants with 7 different features
(e.g., location and price range).
The texts are in the form of conversation
between a user and a bot, in
a restaurant reservation scenario. 
These tasks test the capacity of end-to-end dialog systems with various goals (e.g., request phone-number, address, etc.). Another 
set of test data, named out-of-vocabulary (OOV), are also used to
test the capability of a system to deal with entities not appearing in training data.  

\noindent\textbf{DSTC-2 Dialog datasets.} This restaurant domain dataset \cite{Bordes2016} is similar to bAbI Dialog data in format and goal but contains real human-bot conversations and is derived from second Dialog State Tracking Challenge \cite{DSTC}. This task
draws on human-computer dialog data,
where crowd-workers interacted
with several variants of a spoken dialog system \cite{DSTC}. 
Sample dialog is provided in appendix.
\section{Experimental Results}
\subsection{Travel-Log Results}
We evaluate our model's reasoning capability on 5 different travel log datasets with increased number of attractions.
The action space is composed of the list of attractions and 4 directions.
As a baseline we use \texttt{LSTM}, End-to-End Memory Networks (\texttt{MemN2N})\footnote{\label{n1}Code : github.com/facebook/MemNN \& github.com/facebook/MemNN/tree/master/EntNet-babi} \cite{WestonNIPS2015} and Recurrent Entity Networks
\texttt{Ent-Net}\footref{n1} \cite{WestonICLR17} as summarized in the related work.
We compare these baselines against our full Scaffolding Networks model (\texttt{SN}). 
The importance threshold $I(x_t)$ is empirically chosen to be 0.30. 
To evaluate the impact of the teacher's importance strategy, we trained our model without the importance strategy, where the teacher only uses random sampling to generate questions, (\texttt{SN-no-imp}). To evaluate the importance of our memory attention encoder, we train another \texttt{SN} model without the attention component (\texttt{SN-no-att}). Since teacher's importance strategy depends on the attention, the teacher uses random sampling to generate questions in \texttt{SN-no-att}.  
Table~\ref{atractions} shows benchmark results. Table~\ref{ex-app-log} demonstrates the output of our full \texttt{SN} model on a single log text data. 
We observed the following results:

\begin{itemize}
\item Our full \texttt{SN} models outperform nearly all benchmark models, except for the simplest task.
\vspace{-0.03in}
\item The \texttt{LSTM}s perform poorly. Both \texttt{MemN2N} and \texttt{Ent-Net} performances are similar to our full \texttt{SN} model  for simpler tasks (when \#attractions $\le$10).
\vspace{-0.03in}
\item When no attention is used in \texttt{SN-no-att}, the scaffolding model performance suffers the most. Since only with attention the \texttt{SN} model is able to capture the relations between the previous memory and observed sentence, without attention it's decisions become myopic.
\vspace{-0.03in}
\item With no importance strategy, \texttt{SN-no-imp} performs slightly worse than our full model \texttt{SN}, indicating that teacher's intelligence influences the student's actions. 
\vspace{-0.03in}
\item Although the performance of all models drop as the complexity increase, our \texttt{SN} is both more accurate and more robust on harder tasks. 
\vspace{-0.03in}
\item The \texttt{SN} model can infer locations that doesn't exist in the training data (e.g., north-east) and correctly reason about them (step-17 in Table~\ref{ex-app-log}).
\vspace{-0.03in}
\item The student demonstrates that it can track the state changes and positional relations of the objects
based on the guidance it gets from the teacher (Table~\ref{ex-app-log}). For instance, even though it has not observed the fact about \textit{'west of the park'}, it learns to reason about the answer based on its memory and the scaffolding questions it receives from the teacher in step-16 and step-18, and can successfully answer the corpus provided question at step-20. 
\vspace{-0.03in}
\end{itemize}

\begin{table}[t]
\caption{\small Error Rates on Travel Log test data. Larger number of attractions implies harder tasks.}
\label{atractions}
\vskip -0.15in
\begin{center}
\begin{small}
\begin{tabular}{cccccccc}
\hline
\small Attractions & \small \#Action & \small \texttt{LSTM} & \small \texttt{MemN2N} & \small \texttt{Ent-Net} & \small \texttt{SN}& \small \texttt{SN-no-imp} & \small \texttt{SN-no-att}\\
\hline
\small 5    & \small 9 & \small 41.1 & \small 15.6 & \small 16.6 & \small 18.8 & \small 20.1 & \small 35.0\\
\small 10   & \small 14 & \small 63.6 & \small 41.4 & \small 40.7 & \small \textbf{40.0} & \small 41.1 & \small 59.2\\
\small 15   & \small 19 & \small 88.5 & \small 59.4 & \small 58.4 & \small \textbf{56.3} & \small 58.6 & \small 78.3\\
\small 20   & \small 24 & \small 90.8 & \small 70.4 & \small 69.3 & \small \textbf{67.2} & \small 68.5 & \small 83.9\\
\small 25   & \small 29 & \small 95.5 & \small 79.4 & \small 78.5 & \small \textbf{75.1} & \small 76.2 & \small 89.5\\
\hline
\end{tabular}
\end{small}
\end{center}
\end{table}
\vskip -0.1in
\begin{table}[t]
\renewcommand{\arraystretch}{0.5}
\caption{\small A sample output from scaffolding network inference on travel log. The numbers in parenthesis indicate which sentences the teacher used to generate the questions. The bolded question (20) is corpus provided question.}
\begin{tabular}{@{}m{3.8cm}@{}c@{}m{4.6cm}c@{}@{}r@{}}
\hline
\footnotesize \textbf{Sentences in Travel Log} & \footnotesize \ \ \textbf{Importance} &
\footnotesize \ \  \textbf{Teacher's Question}& \footnotesize \textbf{Student Action} & \footnotesize \ \ \textbf{Reward} \\
\hline
\footnotesize 1: i am at the museum & \footnotesize 0.009 & & & \\
\hline
\footnotesize 2: the school on my east  & \footnotesize 0.015 & \footnotesize (2) \textit{what is on my east} ? & \footnotesize \textit{school} & \footnotesize 1 \\
\hline
\footnotesize 3: i am heading to south & \footnotesize 0.019 & \footnotesize (1) \textit{where am i at} ? & \footnotesize \textit{museum} & \footnotesize 1 \\
\hline
\footnotesize  {4: the coffee-shop on the south of museum} & \footnotesize 0.22 & \footnotesize (4) \textit{what is on the south of the museum} ? & \footnotesize \textit{coffee-shop} & \footnotesize 1 \\
\hline
\footnotesize 5: the parliament is on my north & \footnotesize 0.028 & \footnotesize (1) \textit{where am i at} ? & \footnotesize \textit{museum} & \footnotesize 1 \\
\hline
\footnotesize 6: i am heading south towards train-station & \footnotesize 0.24 & \footnotesize (6) \textit{where am i heading south towards} ? & \footnotesize \textit{train-station} & \scriptsize 1 \\
\hline
\footnotesize 7: after leaving the museum, i reached train-station & \footnotesize 0.47 & \footnotesize (7) \textit{where did i reach ?} & \footnotesize \textit{train-station} & \footnotesize 1 \\
\hline
\footnotesize 8: i am heading east & \footnotesize 0.47 & \footnotesize (2) \textit{what is the school on ?} & \footnotesize \textit{east} & \scriptsize 1 \\
\hline
\footnotesize 9: school is on my north & \footnotesize 0.48 & \footnotesize (2,9) \textit{what is on my east} ? \textit{what is on my north} ? & \footnotesize \textit{school} & \footnotesize 1 \\
\hline
\footnotesize 10: the coffee-shop on my east & \footnotesize 0.53 & \footnotesize (4,10) \textit{what is on the south of the museum} ? \textit{what is on my east} ? & \footnotesize \textit{school} & \footnotesize -1 \\
\hline
\footnotesize 11: the coffee-shop on my east & \footnotesize 0.55 & \footnotesize (10) \textit{what is on my east ?} & \footnotesize \textit{coffee-shop} & \footnotesize 1 \\
\hline
\footnotesize 12: the park is on my east & \footnotesize 0.28 & \footnotesize (1,4) \textit{where am i at} ? \textit{what is the coffee-shop south of} ?& \footnotesize \textit{parliament} & \footnotesize -1 \\
\hline
\footnotesize 13: the park is on my east & \footnotesize 0.57 & \footnotesize (2,13) \textit{what does the school on} ? \textit{what does the park on} ? & \footnotesize \textit{east} & \footnotesize 1 \\
\hline
\footnotesize 14: i am heading to east again & \footnotesize 0.58 & \footnotesize (2) \textit{what is th school on} ? & \footnotesize \textit{east} & \scriptsize 1 \\
\hline
\footnotesize 15: the school on my north & \footnotesize 0.64 & \footnotesize (5,15) \textit{what is parliament on} ? \textit{what is school on} ? & \footnotesize \textit{north} & \footnotesize 1 \\
\hline
\footnotesize 16: i just arrived to the park & \footnotesize 0.68 & \footnotesize (4,7) \textit{what is the coffee-shop south of} ? \textit{where did i leave} ? & \footnotesize \textit{museum} & \footnotesize 1 \\
\hline
\footnotesize 17: i see a restaurant on my south & \footnotesize 0.29 & \footnotesize (2,9) \textit{what is on my east} ? \textit{what is on my north} ? & \footnotesize \textit{school} & \footnotesize 1 \\
\hline
\footnotesize 18: coffee shop is on my west & \footnotesize 0.62 & \footnotesize (4,18) \textit{what is on the south of the museum}? \textit{what is on my west}? & \footnotesize \textit{coffee-shop} & \footnotesize 1 \\
\hline
\footnotesize 19: i am walking towards the restaurant & \footnotesize 0.75 & \footnotesize (17,19) \textit{what do i see on my south ?} \textit{what i am walking towards} ? & \footnotesize \textit{restaurant} & \footnotesize 1 \\
\hline
\footnotesize 20: \textit{\textbf{what is west of the park} ?} & & & \footnotesize \textbf{coffee-shop} & \footnotesize \textit{9}\\
\hline
\end{tabular}
\label{ex-app-log}
\vskip -0.15in
\end{table}
\subsection{Dialog Results on bAbI and DSTC-2 Datasets}
The input text in both dialog data is defined as a series of utterances $c^u_1,c^s_1,c^u_2,c^s_2,..,c^u_{t-1},c^s_{t-1}$ (alternating between the user utterance $c^u_i$ and agent's response $c^s_i$), while the $c^u_t$ is the question and the goal is to predict agent's response $c^s_t$. 
In this task the $c^s_t$ is a sequence of words, while our model predicts word tokens as actions. Therefore, we map each response $c^s_t$ to a unique category following \cite{Bordes2016}. Specifically, we collect all possible agent responses into a candidate set $\mathcal{C}$ and define an action $a_i$ as the i$^{th}$ response in the candidate set $\mathcal{C}$ such that $a_i\in \mathcal{C}$. 
The set of candidate responses includes all possible bot utterances and API calls.
We benchmark our approach against: End-to-End Goal Oriented Dialog Model (\texttt{N2N}) \cite{Bordes2016}, Query Reduction Networks (\texttt{QRN}) \cite{Seo},
and Gated Memory Networks (\texttt{GMemN2N}) \cite{GE2EMN}, all presented state-of-the art in this dataset.
We show the results of our full \texttt{SN}s with teacher's importance strategy. 
Performance results are summarized in Table~\ref{babi-dialog} comparing models when no-match and 7 different restaurant specific match features are used, e.g., location, price range, etc., indicating if there is an exact match between words occurring in the current utterance and those in the question or memory. We observed the following results:
\begin{itemize}
\item For Tasks 1-5, when no features used, our \texttt{SN} model follows the state-of-the-art closely.
\vspace{-0.05in}
\item Especially when match features are used, on average \texttt{SN} outperforms all models. 
\vspace{-0.05in}
\item The OOV experiments in Table~\ref{babi-dialog} demonstrate how the models handle actions with "unseen" entities in conversational dialogs. All models have slightly higher error rates with OOV words. \texttt{SN}s immediately follow the best model when no match features (plain) are used, while outperforms all others when match features are used.   
\vspace{-0.05in}
\item The efficacy of our \texttt{SN} models are more pronounced with and without the match features with the real dialog scenarios in DSTC-2.
\end{itemize}

\begin{table}[t]
\vskip -0.3in
\caption{\small Error Rates on Babi Dialog and DSTC2 dialog datasets \cite{Bordes2016} with and without match features.}
\label{babi-dialog}
\vskip -0.3in
\begin{center}
\begin{small}
\begin{tabular}{@{}l@{}c@{}|@{}c@{}c@{}c@{}c|c@{}c@{}c@{}c@{}}
\hline
& \small \#Action \ & \multicolumn{4}{|c|}{\small No-Match Features} &  \multicolumn{4}{c}{\small With Match Features} \\
\hline
\small \ Task \ \ & \small $|\mathcal{C}|$ \ \mbox& \small  \ \ \texttt{N2N} \ \ & \small \texttt{QRN} & \ \small \texttt{GMemN2N} \ & \small \texttt{SN}  &\small  \texttt{N2N} \ \ & \small \texttt{QRN} & \ \small \texttt{GMemN2N} \ &\small \texttt{SN}\\
\hline
\footnotesize 1: Issuing API Calls      & \small 309 & \small 0.1 & \small \textbf{0.0} & \small \textbf{0.0} & \small \textbf{0.0} & \small \textbf{0.0} & \small \textbf{0.0} & \small \textbf{0.0} & \small \textbf{0.0}\\
\footnotesize 2: Updating API Calls    & \small 307  & \small \textbf{0.0} & \small \textbf{0.0}& \small \textbf{0.0}& \small \textbf{0.0} & \small 1.7 & \small \textbf{0.0} & \small \textbf{0.0}& \small \textbf{0.0}  \\
\footnotesize 3: Displaying Options   & \small 333   & \small 25.1 & \small 12.6 & \small 25.1 &\small  \textbf{12.0}  & \small 25.1 & \small 7.6 & \small 25.1& \small \textbf{5.6}\\
\footnotesize 4: Providing Extra Information  & \small 1036 & \small 40.5 & \small \textbf{14.3} & \small 42.8 & \small \textbf{14.2} & \small 0.0 & \small 0.0  & \small 0.0 & \small 0.0\\ 
\footnotesize 5: Conducting Full Dialogs   & \small 1257 & \small 3.9 & \small \textbf{0.7} & \small 3.7 & \small 3.4 & \small 6.6 & \small \textbf{0.0} & \small  2.0 & \small 1.0  \\ 
\hline
\footnotesize Average Error Rates (\%) & & \small 13.9 & \small \textbf{5.5} & \small 14.3 & \small 5.9 & \small 6.7 & \small 1.5 & \small 5.4 & \small \textbf{1.3} \\
\hline
\footnotesize 1: (OOV) Issuing API Calls  & \small 309    & \small 27.7 & \small 6.6 & \small 17.6& \small \textbf{5.6}  & \small 3.5 &\small \textbf{0.0} & \small \textbf{0.0} & \small 1.2 \\
\footnotesize 2: (OOV) Updating API Calls & \small 307    & \small 21.1 & \small \textbf{8.4} & \small 21.1 &\small  10.8 & \small 5.5 &\small  \textbf{0.0} & \small 5.8 & \small 0.8\\
\footnotesize 3: (OOV) Displaying Options  & \small 333   & \small 25.6 & \small 12.4 & \small 24.7 &\small  \textbf{10.2} & \small 24.8 & \small 7.7  & \small 24.9 & \small \textbf{6.0}\\
\footnotesize 4: (OOV) Providing Extra Inform. & \small 1036  & \small  42.4 & \small \textbf{14.4}& \small 43.0 & \small 23.3&\small  \textbf{0.0} & \small  \textbf{0.0} & \small \textbf{0.0} & \small \textbf{0.0}  \\ 
\footnotesize 5: (OOV) Conducting Full Dialogs & \small 1257  & \small 34.5 & \small \textbf{13.7} & \small 33.3 & \small 14.6 & \small 22.3 &\small 4.0 & \small 20.6 & \small \textbf{2.6} \\ 
\hline
\footnotesize Average Error Rates & & \small 30.3 & \small \textbf{11.1} & \small 26.9 &\small  12.9 & \small 11.2 & \small 2.3 & \small 10.3 & \small \textbf{2.1}\\
\hline
\footnotesize DSTC-2 Real Dialogs  & \small 1257 & \small 58.9 & \small 48.9 & \small 52.6 & \small \textbf{47.6}& \small 59.0 & \small 49.3  & \small 51.3 & \small \textbf{48.7} \\
\hline
\end{tabular}
\end{small}
\end{center}
\vskip -0.1in
\end{table}
\clearpage\pagebreak
\noindent\textbf{Scaffolding Teacher versus the Human as a Teacher. }
\label{exp-self}
We investigate how much improvement we gain from the teacher generated questions by gradually replacing the human provided questions with
\newcommand{\demofig}
\noindent\demofig
\begin{wrapfigure}{r}{0.34\textwidth}
\vskip -0.3in
\begin{center} 
\begin{tikzpicture}[scale=0.5,font=\large]
\begin{axis}[title=\textbf{DSTC-2 Dialog Data},xlabel=\%human generated data used,
xtick={10,25,50,75,100},
grid=both,grid style={black!15},
ylabel=Test Error (\%),legend pos=north east]
\addplot coordinates {
(10, 73)
(25, 64)
(50, 58)
(75, 53)
(100, 47.6)
};
\addplot coordinates {
(10.0, 78)
(25, 70)
(50, 63)
(75, 55)
(100, 48.9)
};
\legend{\texttt{SN},\texttt{QRN}}
\end{axis}
\end{tikzpicture}
\end{center} 
\vskip -0.15in
\caption{\small Error rates showcasing the efficacy of the models when the number of human input is reduced.}
\label{selfsupervised}
\vskip -0.15in
\end{wrapfigure}
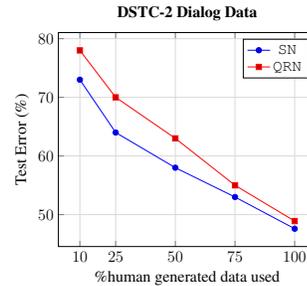
the teacher's questions at training time. We benchmark \texttt{SN} against \texttt{QRN}~\cite{Seo} (best performing benchmark). Specifically, for \texttt{SN} models, we only use \textit{m}\% of the human generated questions and replace the rest with the teacher generated ones and re-train.  
For \texttt{QRN} models, we use self-training by first removing m\% of the training data, and training with the rest (100-\textit{m})\% human labeled data. Then we predict the answers for the removed \textit{m}\% text, add them back to the training data, re-train and repeat the process. We repeat the experiments for increasing values of $m$ and demonstrate the results on DSTC-2 real dialog dataset. 

As shown in in Figure~\ref{selfsupervised}, reducing the human provided input does not affect the \texttt{SN}s performance as much as \texttt{QRN} models (e.g., 5\% reduction in error with the \texttt{SN} models when 50\% of the labels are removed).
Even with reduced human input, the \texttt{SN}s can learn to do reasoning well and track the state of the world, reducing the dependence
on humans as teachers. 
\section{Conclusion and Outlook}
\label{future}
We take a step towards teaching machines to reason by introducing an RL framework that simulates instructional scaffolding teaching. We found that by integrating the strengths of neural attention models and incremental teacher-student learning we can repeatably check and update the student about the information it encodes in sequential order. 
We additionally show that with a simple importance strategy the teacher can guide the student into learning a policy for identifying not only new aspects but also track their state changes throughout the text. 
Our future work will investigate a teacher acting as another agent to influence the student into taking actions. 
\clearpage
\pagebreak
\clearpage
\pagebreak
\bibliography{SNMain}

\begin{thebibliography}{}

\bibitem[\protect\citename{Andreas {\em et~al.}, }2016]{Andreas}
Andreas, Jacob, Rohrbach, Marcus, Darrell, Trevor, \& Klein, Dan. 2016.
\newblock Learning to compose neural networks for question answering.
\newblock {\em In:} {\em NAACL}.

\bibitem[\protect\citename{Bordes \& Weston, }2017]{Bordes2016}
Bordes, Antoine, \& Weston, Jason. 2017.
\newblock Learning end-to-end goal-oriented dialog.
\newblock {\em In:} {\em ICLR}.

\bibitem[\protect\citename{Boyd-Graber {\em et~al.}, }2015]{Graber}
Boyd-Graber, Jordan, Iyye, Mohit, He, He, \& III, Hal~Daumé. 2015.
\newblock Interactive Incremental Question Answering.
\newblock {\em In:} {\em NIPS}.

\bibitem[\protect\citename{Guo {\em et~al.}, }2017]{GuoICLR17}
Guo, Xiaoxiao, Klinger, Tim, Rosenbaum, Clemens, Bigus, Joseph~P., Campbell,
  Murray, Kawass, Ban, Talamadupula, Kartik, \& Tesauro, Greald. 2017.
\newblock Learning to Query, Reason and Answer Question On Ambiguous Texts.
\newblock {\em In:} {\em ICLR}.

\bibitem[\protect\citename{Heilman \& Smith, }2010]{Heilman}
Heilman, Michael, \& Smith, Noah~A. 2010.
\newblock Good Question! Statistical Ranking for Question Generation.
\newblock {\em In:} {\em NAACL}.

\bibitem[\protect\citename{Henaff {\em et~al.}, }2017]{WestonICLR17}
Henaff, Mikhael, Weston, Jason, Szlam, Arthur, Bordes, Antoine, \& LeCun, Yann.
  2017.
\newblock Tracking the World State With Recurrent Entity Networks.
\newblock {\em In:} {\em ICLR}.

\bibitem[\protect\citename{Henderson {\em et~al.}, }2014]{DSTC}
Henderson, Matt, Thomson, Blaise, \& Williams, Jason. 2014.
\newblock The second dialog state tracking challenge.
\newblock {\em In:} {\em Annual Meeting of the Special Interest Group on
  Discourse and Dialogue}.

\bibitem[\protect\citename{Hochreiter \& Schmidhuber, }1997]{Hochreiter}
Hochreiter, Sepp, \& Schmidhuber, Jurgen. 1997.
\newblock Long short-term memory.
\newblock {\em Neural Computation}, {\bf 9}(8), 1735--1780.

\bibitem[\protect\citename{Hogan \& Pressley, }1997]{Hogan}
Hogan, K., \& Pressley, M. 1997.
\newblock {\em Scaffolding student learning: Instructional approaches and
  issues}.
\newblock Cambridge, MA: Brookline Books.

\bibitem[\protect\citename{Kim, }2010]{Youb}
Kim, Youb. 2010.
\newblock Scaffolding Through Questions in Upper Elementary ESL Learning.
\newblock {\em Literacy Teaching and Learning}, {\bf 15}(1--2), 109--137.

\bibitem[\protect\citename{Kingma \& Ba, }2015]{Kigma}
Kingma, Diederik~P., \& Ba, Jimmy. 2015.
\newblock Adam: A method for stochastic optimization.
\newblock {\em In:} {\em 3rd International Conference for Learning
  Representations, San Diego}.

\bibitem[\protect\citename{Klein \& Manning, }2003]{StanfordParser}
Klein, Dan, \& Manning, Christopher. 2003.
\newblock Fast exact inference with a factored model for natural language
  parsing.
\newblock {\em In:} {\em NIPS}.

\bibitem[\protect\citename{Levy \& Andrew, }2006]{Tregex}
Levy, R., \& Andrew, G. 2006.
\newblock Tregex and Tsurgeon: tools for querying and manipulating tree data
  structures.
\newblock {\em In:} {\em LREC}.

\bibitem[\protect\citename{Li {\em et~al.}, }2017]{Jiwei}
Li, Jiwei, Miller, Alexander~H., Chopra, Sumit, Ranzato, Marc’Aurelio, \&
  Weston, Jason. 2017.
\newblock Dialog Learning with Human-in-the-loop.
\newblock {\em In:} {\em ICLR}.

\bibitem[\protect\citename{Mnih {\em et~al.}, }2015]{Mnih2015}
Mnih, Volodymyr, Kavukcuoglu, Koray, Silver, David, Rusu, Andrei~A., Veness,
  Joel, Bellemare, Marc~G., Graves, Alex, Riedmiller, Martin, Fidjeland,
  Andreas~K., Ostrovski, Georg, Petersen, Stig, Beattie, Charles, Sadik, Amir,
  Antonoglou, Ioannis, King, Helen, Kumaran, Dharshan, Wierstra, Daan, Legg,
  Shane, \& Hassabis, Demis. 2015.
\newblock Human-level control through deep reinforcement learning.
\newblock {\em Nature}, {\bf 518}, 529--533.

\bibitem[\protect\citename{Perez \& Liu, }2016]{GE2EMN}
Perez, Julien, \& Liu, Fei. 2016.
\newblock Gated End-to-End Memory Networks.
\newblock {\em In:} {\em arxiv pre-print arXiv:1610.04211}.

\bibitem[\protect\citename{Rocktaschel {\em et~al.}, }2016]{Tim}
Rocktaschel, Tim, Grefenstette, Edward, Hermann, Karl~Moritz, Kocisky, Tomas,
  \& Blunsom, Phil. 2016.
\newblock Reasoning about Entailment with Neural Attention.
\newblock {\em In:} {\em ICLR}.

\bibitem[\protect\citename{Seo {\em et~al.}, }2017]{Seo}
Seo, Minjoon, Min, Sewon, Farhadi, Ali, \& Hajishirzi, Hannaneh. 2017.
\newblock Query-Reduction Networks for Question Answering.
\newblock {\em In:} {\em ICLR}.

\bibitem[\protect\citename{Shen {\em et~al.}, }2016]{Yelong}
Shen, Yelong, Huang, Po-Sen, Gao, Jianfeng, \& Chen, Weizhu. 2016.
\newblock ReasoNet: Learning to Stop Reading in Machine Comprehension.
\newblock {\em In:} {\em arXiv Preprint arXiv:1611.04642}.

\bibitem[\protect\citename{Suay \& Chernova, }2011]{Suay}
Suay, H.B., \& Chernova, S. 2011.
\newblock Effect of human guidance and state space size on Interactive
  Reinforcement Learning.
\newblock {\em In:} {\em RO-MAN, IEEE}.

\bibitem[\protect\citename{Sukhbaatar {\em et~al.}, }2015]{WestonNIPS2015}
Sukhbaatar, Sainbayar, Szlam, Arthur, Weston, Jason, \& Fergus, Rob. 2015.
\newblock End-To-End Memory Networks.
\newblock {\em In:} {\em NIPS}.

\bibitem[\protect\citename{Wang {\em et~al.}, }2016]{Wang16}
Wang, Sida~I., Liang, Percy, \& Manning, Christopher~D. 2016.
\newblock Learning language games through interaction.
\newblock {\em In:} {\em ACL}.

\bibitem[\protect\citename{Weston, }2016]{WestonDialog}
Weston, Jason. 2016.
\newblock Dialog Based Language Learning.
\newblock {\em In:} {\em NIPS}.

\bibitem[\protect\citename{Yu {\em et~al.}, }2017]{Baidu}
Yu, Haonan, Zhang, Haichao, \& Xu, Wei. 2017.
\newblock A Deep Compositional Framework for Human-like Language Acquisition in
  Virtual Environment.
\newblock {\em In:} {\em arXiv preprint arXiv:1703.09831}.

\bibitem[\protect\citename{Zimmer {\em et~al.}, }2014]{Zimmer}
Zimmer, Matthieu, Viappiani, Paolo, \& Weng, Paul. 2014.
\newblock Teacher-Student Framework: A Reinforcement Learning Approach.
\newblock {\em In:} {\em AAMAS Workshop Autonomous Robots and Multirobot
  Systems}.

\bibitem[\protect\citename{Zubek \& Dietterich, }2002]{Zubek}
Zubek, Valentina~Bayer, \& Dietterich, Thomas~G. 2002.
\newblock Pruning improves heuristic search for cost-sensitive learning.
\newblock {\em In:} {\em ICML}.

\end{thebibliography}
\bibliographystyle{authordate1}

\clearpage
\pagebreak

 \section*{Appendix}
 \label{appendix}
 \subsection*{Appendix-A : Details of the Travel Log Experiments}
 \label{appexb}
 The traveler is placed on a 9$\times$9 grid town of 81 distinct locations with a list of attractions in randomly chosen locations.
 At each time step, the traveler's next direction is chosen at random among 4 possible actions, e.g., north, south, east, west. If the selection puts the traveler off the grid, a new legal direction is chosen at random. 

 We use the following rules to generate the log data. 
 The traveler visits each attraction only once. If the randomly chosen next direction takes the traveler to an attraction previously visited, a new random direction is chosen. If all the 4 directions contain attractions that are visited before, the log terminates. Once the traveler is in a new location, she looks at 4 immediate directions to log attractions. If there is no attraction, she continues on the next direction. Otherwise randomly chooses one attraction to log.   
 We randomly generate different town layouts by varying the number of attractions, e.g., $\{$\textit{5,10,15,20,25}$\}$ and evaluate the performance of the models in the experiments. The teacher only uses the sentences that contain the location of the attractions to generate questions.

 We use all the words in the vocabulary except a list of stop words. 
 The list of attractions and 4 different directions are used as the output space. 
 If the attraction is a multi-word expression, we combine all the words.
 \subsection*{Appendix-B : Example Dialog from the DSTC-2 Dataset}
 \label{appexc}
 \begin{center}
 \begin{tabular}{cp{7cm}}
 \hline
 Turn & User Utterance (\textbf{U}) / System Response (\textbf{S})\\
 \hline
 \small 1 & \small (\textbf{U}) <SILENCE> \\ & \small (\textbf{S}) Hello , welcome to ...  How may I help you ? \\
 \small 2 & \small (\textbf{U}) i want a moderately priced restaurant that serves \\ &  \small (\textbf{S}) What kind of food would you like? \\
 \small 3 &  \small (\textbf{U}) canapes food \\ & \small (\textbf{S}) What part of town do you have in mind? \\
 \small 4 & \small (\textbf{U}) it doesnt matter \\ & \small (\textbf{S}) \textit{api-call} canapes \textit{R-location} moderate \\
 \small 5 & \small \textit{api-call} no result \\
 \small 6 & \small (\textbf{U}) <SILENCE> \\ &\small  (\textbf{S}) \textit{yippee\_noodle\_bar} is a nice restaurant in the centre of town in the moderate price range \\
 \small 7 & \small (\textbf{U}) gastropub food \\ & \small (\textbf{S}) Sorry there is no canapes restaurant in the moderate price range \\
 \small 8 & \small (\textbf{U}) i want a gastropub restaurant \\ & \small (\textbf{S}) the cow pizza kitchen and bar is a nice restaurant in the centre of town serving gastropub food \\
 \small 9 &\small  (\textbf{U}) what is the address \\ & \small (\textbf{S}) Sure , the cow pizza kitchen and bar is on  \textit{the-cow-pizza-kitchen-and-bar-address} \\
 \small 10 & \small (\textbf{U}) what is the phone number \\& 	\small (\textbf{S}) The phone number of the cow pizza kitchen and bar is \textit{the-cow-pizza-kitchen-and-bar-phone} \\
 \small 11 &\small  (\textbf{U}) what is the area \\ & \small (\textbf{S}) the cow pizza kitchen and bar is a nice place in the centre of town \\
 \small 12 & \small (\textbf{U}) thank you goodbye \\ & \small (\textbf{S}) you are welcome \\
 \hline
 \end{tabular}
 \end{center}
\end{document}